\documentclass{article}

\usepackage{microtype}
\usepackage{graphicx}
\usepackage{subfigure}
\usepackage{etoolbox}
\usepackage{tabularx}
\usepackage{color,soul}
\usepackage{booktabs} 

\usepackage{hyperref}

\usepackage[accepted]{icml2023}

\usepackage{amsmath}
\usepackage{amssymb}
\usepackage{mathtools}
\usepackage{amsthm}

\usepackage[capitalize,noabbrev]{cleveref}

\theoremstyle{plain}

\theoremstyle{definition}

\theoremstyle{remark}

\usepackage[textsize=tiny]{todonotes}

\icmltitlerunning{Gradient-free training of neural ODEs for system identification and control}

\def\ie{\textit{i.e.}}

\begin{document}
\twocolumn[
\icmltitle{Gradient-free training of neural ODEs for system identification and control \\ using ensemble Kalman inversion}



\icmlsetsymbol{equal}{*}

\begin{icmlauthorlist}
\icmlauthor{Lucas B\"ottcher}{yyy}
\end{icmlauthorlist}

\icmlaffiliation{yyy}{Department of Computational Science and Philosophy, Frankfurt School of Finance and Management, Frankfurt am Main, Germany}

\icmlcorrespondingauthor{Lucas B\"ottcher}{l.boettcher@fs.de}

\icmlkeywords{dynamical systems, data-driven dynamics, neural ODEs, optimal control, ensemble Kalman inversion}

\vskip 0.3in
]



\printAffiliationsAndNotice{}  

\begin{abstract}
Ensemble Kalman inversion (EKI) is a sequential Monte Carlo method used to solve inverse problems within a Bayesian framework. Unlike backpropagation, EKI is a gradient-free optimization method that only necessitates the evaluation of artificial neural networks in forward passes. In this study, we examine the effectiveness of EKI in training neural ordinary differential equations (neural ODEs) for system identification and control tasks. To apply EKI to optimal control problems, we formulate inverse problems that incorporate a Tikhonov-type regularization term. Our numerical results demonstrate that EKI is an efficient method for training neural ODEs in system identification and optimal control problems, with runtime and quality of solutions that are competitive with commonly used gradient-based optimizers.
\end{abstract}

\section{Introduction}
Already in 1988, two years after \citet{rumelhart1986learning} have proposed backpropagation as a training method for multilayer perceptrons, \citet{singhal1988training} used an extended Kalman filter (EKF) to train such neural networks. In a follow-up study, \citet{singhal1989training} noted that ``the Kalman
algorithm converges in fewer iterations than back-propagation and obtains solutions with fewer hidden nodes in the network.'' While a more detailed comparison of backpropagation and EKFs showed that the latter may be associated with a substantially higher computational cost~\cite{ruck1992comparative}, a more efficient variant of the EKF used by Singhal and Wu has been introduced by~\citet{puskorius1991decoupled} and been shown to converge towards desired solutions more rapidly than backpropagation in different learning tasks. Until 2019, when \citet{kovachki2019ensemble} used a Monte Carlo approximation of the EKF, the so-called ensemble Kalman filter (EnKF)~\cite{evensen1994sequential}, to train artificial neural networks (ANNs), most applications of Kalman filters to such training tasks were based on extended and unscented Kalman filters~\cite{haykin2001kalman} that are known to suffer from computational and memory limitations. The EnKF avoids these limitations by propagating an ensemble of states that approximates the system state distribution and from which covariance matrix estimates are computed at every iteration. This method originated in the geosciences and has been successfully applied to many high-dimensional and non-linear data assimilation problems~\cite{katzfuss2016understanding}. In addition to its application in data assimilation, the EnKF has been adapted to solve general inverse problems of the form
\begin{equation}
    y = G(\theta)+\xi\,,
    \label{eq:inverse_problem}
\end{equation}
where one wishes to determine model parameters $\theta\in\mathcal{U}$ based on a known output variable $y\in\mathcal{Y}$ and model $G\colon \mathcal{U}\rightarrow \mathcal{Y}$~\cite{iglesias2013ensemble}. The quantity ${\xi \sim \mathcal{N}(0,\Gamma)}$ denotes Gaussian noise with covariance $\Gamma$. The use of EnKF iterations to solve inverse problems has been dubbed ``ensemble Kalman inversion'' (EKI). A continuous-time limit of the discrete-time formulation of EKI has been derived by \citet{schillings2017analysis}. 

Ensemble Kalman inversion belongs to the class of sequential Monte Carlo methods that are used to solve inverse problems within a Bayesian framework~\cite{idier2013bayesian,Dashti2017}. \citet{kovachki2019ensemble} have cast supervised, semi-supervised, and online learning tasks into inverse problems of the form (\ref{eq:inverse_problem}) and solved them using EKI. Unlike backpropagation, EKI is a gradient-free optimization method that only requires one to evaluate ANNs in forward passes. It can thus be easily parallelized. 

Another variant of the EnKF has been used by \citet{haber2018never} to solve non-linear regression and image classification problems and an EKI-based sparse learning method has been proposed by \citet{schneider2022ensemble} to use time-averaged statistics for data-driven discovery of differential equations. Ensemble Kalman methods have also been combined with auto-differentation approaches for parameter inference in dynamical systems and neural network models of partially or fully unknown dynamics~\cite{chen2022autodifferentiable}.

In this work, we study the ability of EKI to efficiently train neural ordinary differential equations (neural ODEs) in system identification and control tasks. Neural ODEs have received renewed interest in recent years due to their diverse applications in dynamical systems identification, timeseries modeling~\cite{wang1998runge,chen2018neural} and optimal control~\cite{asikis2022neural,bottcher2022ai,bottcher2022near,cuchiero2020deep}. Inverse problems (\ref{eq:inverse_problem}) can be generalized to describe optimal control problems that involve an additional Tikhonov-type regularization term~\cite{clason2020optimal}. Complementing earlier work on Tikhonov EKI~\cite{chada2020tikhonov}, we incorporate a regularization term in equation (\ref{eq:inverse_problem}) to solve optimal control problems with EKI and neural ODEs.
\section{Contributions}
The main contributions of our work are as follows:
\begin{itemize}
    \item Combining EKI with neural ODEs for system identification tasks.
    \item Formulating optimal control problems in terms of an inverse problem (\ref{eq:inverse_problem}) that accounts for a control-energy regularization term.
    \item Formulating EKI updates for solving optimal control problems with neural ODEs.
    \item Comparing gradient-based and EKI-based optimization of neural ODEs in system identification and control tasks.
\end{itemize}
Our source codes are publicly available at \cite{gitlab}.
\section{Ensemble Kalman inversion}
\label{sec:eki}
Ensemble Kalman inversion is a gradient-free optimization method that can be used to solve inverse problems (\ref{eq:inverse_problem}) in an iterative manner. Because different neural-network optimization problems can be cast in the form (\ref{eq:inverse_problem})~\cite{kovachki2019ensemble}, we will use EKI in this work to determine the optimal neural ODE parameters $\theta^*\in\mathcal{U}$ that minimize a given loss function. In Sections~\ref{sec:learning_dyn_sys} and \ref{sec:optimal_control}, we will reframe neural ODE-based system identification and optimal control problems as inverse problems and demonstrate how they can be solved using EKI. We compare the ability of EKI to efficiently identify solutions that are close to the desired optimum with algorithms that rely on backpropagation through time (BPTT)~\cite{williams1990efficient,werbos1990backpropagation,feldkamp1993neural}.

Based on a discrete-time formulation of the EnKF for inverse problems~\cite{iglesias2013ensemble}, a corresponding continuous-time formulation has been derived by \citet{schillings2017analysis}. As a starting point, we consider an ensemble $\{\theta^{(j)}\}_{j=1}^J \subset \mathcal{U}$ of neural-network parameters. In accordance with \citet{schillings2017analysis}, the continuous-time evolution of the ensemble $\{\theta^{(j)}\}_{j=1}^J$ is described by
\begin{align}
    \dot{\theta}^{(j)}&=-C^{\theta G}(\theta) \Gamma^{-1} \left(G(\theta^{(j)})-y\right)\,,\label{eq:dthetadt}\\
    \theta^{(j)}(0)&=\theta_0^{(j)}\,,
\end{align}
where the empirical cross-covariance matrix is
\begin{equation}
    C^{\theta G}(\theta)=\frac{1}{J}\sum_{j=1}^J \left(\theta^{(j)}-\bar{\theta}\right)\otimes \left(G(\theta^{(j)})-\bar{G}\right)\,.
\end{equation}
The ensemble means $\bar{\theta}$ and $\bar{G}$ are given by
\begin{equation}
    \bar{\theta}=\frac{1}{J}\sum_{j=1}^J \theta^{(j)}\quad\mathrm{and}\quad\bar{G}=\frac{1}{J}\sum_{j=1}^J G(\theta^{(j)})\,,
    \label{eq:bar_theta}
\end{equation}
respectively.

For linear inverse problems where $G(\theta)=A\theta$, it has been shown by \citet{schillings2017analysis} that Eq.~(\ref{eq:dthetadt}) can be rewritten as
\begin{equation}
    \dot{\theta}^{(j)}(t)=-C(\theta)\nabla_\theta\Phi(\theta;y)\,,
\label{eq:linear_EKI}
\end{equation}
where
\begin{equation}
    \Phi(\theta;y)=\frac{1}{2}\|y-A\theta\|_\Gamma^2\equiv \frac{1}{2}\| \Gamma^{-1/2} (y-A \theta)\|_{\mathcal{Y}}^2
\end{equation}
and
\begin{equation}
    C(\theta)=\frac{1}{J}\sum_{j=1}^J \left(\theta^{(j)}-\bar{\theta}\right)\otimes \left(\theta^{(j)}-\bar{\theta}\right)\,.
\end{equation}
For any symmetric, positive-definite operator ${C\colon \mathcal{H}\rightarrow\mathcal{H}}$, we use the notations ${\|\cdot\|_C=\|C^{-1/2}\|_{\mathcal{H}}}$ and ${\langle \cdot, \cdot \rangle_C = \langle C^{-1/2} \cdot, C^{-1/2} \cdot \rangle}_{\mathcal{H}}$ to indicate adapted versions of the norm $\|\cdot\|_{\mathcal{H}}$ and inner product $\langle \cdot, \cdot \rangle_{\mathcal{H}}$ associated with a Hilbert space $\mathcal{H}$. 

Equation~(\ref{eq:linear_EKI}) shows that ${\theta^{(j)}\rightarrow\theta^*}$ in the limit ${t\rightarrow\infty}$ where $\theta^*$ minimizes $\Phi(\theta;y)$ within the subspace ${\mathrm{span}\{\theta_0^{(j)}-\bar{\theta}_0\}_{j=1}^J}$ of $\mathcal{U}$. Here, $\bar{\theta}_0$ is the mean of the initial ensemble $\{\theta_0^{(j)}\}_{j=1}^J$. 
\section{Learning dynamical systems}
\label{sec:learning_dyn_sys}
\subsection{Neural ODEs}
\label{sec:neural_odes}
A neural ODE parameterizes the vector field ${f\colon \mathbb{R}^n\times  \mathbb{R} \rightarrow \mathbb{R}^n}$ of a dynamical system
\begin{equation}
    \dot{x}(t)=f(x(t),t)\,,
    \label{eq:dyn_sys_node}
\end{equation}
where $x(t)\in \mathbb{R}^n$ is the system state at time $t$. We use $x(0)=x_0\in\mathbb{R}^n$ to denote the corresponding initial condition. 

Specifically, a neural ODE is an artificial neural network $f_\theta(x(t),t)$ with parameters $\theta\in\mathbb{R}^N$ that is used to represent the right-hand side of Eq.~(\ref{eq:dyn_sys_node}). When numerically integrating the dynamical system (\ref{eq:dyn_sys_node}) with vector field $f_\theta(x(t),t)$, we say that the underlying artificial neural network becomes \emph{time-unfolded}. For example, considering a simple forward Euler scheme with time step $\Delta t$, we have
\begin{equation}
    x_{k+1}=x_k+\Delta t f_\theta(x_k,t_k)\,,
    \label{eq:time_unfolding}
\end{equation}
where $x_k\equiv x(t_k)$ and $t_k=k\Delta t$ for some positive integer $k$. Equation~(\ref{eq:time_unfolding}) shows that integrating the dynamical system (\ref{eq:dyn_sys_node}) produces a residual neural network~\cite{DBLP:conf/cvpr/HeZRS16} that satisfies 
\begin{equation}
    f_\theta(x_k,t_k)=f_\theta(x_{k-1}+\Delta tf_\theta(x_{k-1},t_{k-1}),t_k)\,.
\end{equation}

Given observations $\{\hat{x}(t_\ell)\}_{\ell\in\{1,\dots,M\}}$ at times $t_\ell\in[0,T]$, our goal is to learn $f_\theta(x(t),t)$. To do so, we use the mean squared error
\begin{equation}
    {\rm MSE}(\theta) = \frac{1}{M}\sum_{\ell=1}^M\left[\hat{x}(t_\ell)-x(t_\ell;\theta)\right]^2
    \label{eq:mse_loss}
\end{equation}
as a loss function and train a neural ODE with a suitable optimizer that minimizes Eq.~(\ref{eq:mse_loss}).
\subsection{Numerical experiments}
\begin{figure}[ht!]
\vskip 0.2in
\begin{center}
\centerline{\includegraphics{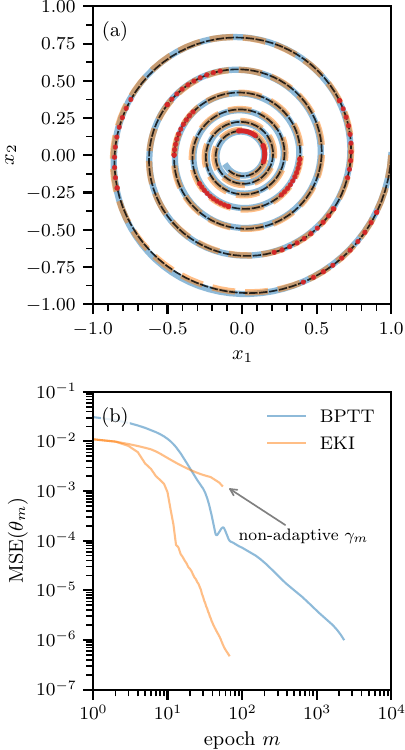}}
\caption{Learning a dynamical system from observation data using BPTT and EKI. (a) We employ a neural ODE to learn the dynamical system (\ref{eq:2d_example}). Solid blue and orange lines correspond to solutions that are obtained with BPTT and EKI, respectively. The dashed black line indicates the solution of the original dynamical system for $(x_1(0),x_2(0))^\top=(1,0)^\top$. Red dots represent observation data that is used to train the neural ODE. (b) Mean squared error (MSE) associated with BPTT (solid blue line) and EKI (solid orange line) as a function of the number of training epochs $m$ [see Eq.~(\ref{eq:mse_loss})]. The neural ODE consists of one hidden layer with 10 $\tanh$ neurons. Its total number of parameters is 52. In the gradient-based optimization, we used the Adam optimizer and set the learning rate to $\eta=0.01$. The EKI optimizer has 22 ensemble members and an exponential scheduler to adapt $\gamma_m$ [see Eq.~(\ref{eq:exp_schedule})]. For EKI, we also show the MSE evolution if no scheduler is used for $\gamma_{m}$. The EKI MSE is based on the minimum MSE in the whole ensemble. We stopped both training algorithms when their runtime exceeded one minute.}
\label{fig:neural_ode_spiral}
\end{center}
\vskip -0.2in
\end{figure}
\begin{table*}[ht!]
\centering
\renewcommand*{\arraystretch}{1.6}
\small
\begin{tabular}{
>{\raggedright\arraybackslash} m{7em}
>{\raggedright\arraybackslash} m{5em}
>{\raggedright\arraybackslash} m{5em}
>{\raggedright\arraybackslash} m{5em}
>{\raggedright\arraybackslash} m{5em}
>{\raggedright\arraybackslash} m{5em}
}\toprule
& EKI & SGD ($\eta=0.01$) & SGD ($\eta=0.1$) & Adam ($\eta=0.01$) & Adam ($\eta=0.1$) \\[1pt] \hline
training error & $\mathbf{4.89\times 10^{-7}}$ & $2.47\times 10^{-3}$ & $5.02\times 10^{-5}$ & $1.03\times 10^{-6}$ & $5.79\times 10^{-7}$\\[1pt]
test error & $\mathbf{9.11\times 10^{-4}}$ & $2.48\times 10^{-1}$ & $5.47\times 10^{-3}$ & $1.12\times 10^{-3}$ & $7.86\times 10^{-3}$\\[1pt]
 \bottomrule
\end{tabular}
\vspace{1mm}
\caption{Training and test errors associated with different optimizers. Training has been stopped after the training time exceeded 60 seconds. SGD and Adam results are based on the same initial neural-network parameters. For EKI, the initial ensemble size is 22 and the diagonal elements of the covariance matrix are updated according to the exponential scheduler (\ref{eq:exp_schedule}).}
\label{tab:spiral_performance}
\end{table*}
\begin{table*}[ht!]
\centering
\renewcommand*{\arraystretch}{1.6}
\small
\begin{tabular}{
>{\raggedright\arraybackslash} m{7em}
>{\raggedright\arraybackslash} m{5em}
>{\raggedright\arraybackslash} m{5em}
>{\raggedright\arraybackslash} m{5em}
>{\raggedright\arraybackslash} m{5em}
>{\raggedright\arraybackslash} m{5em}
}\toprule
& EKI & SGD ($\eta=0.01$) & SGD ($\eta=0.1$) & Adam ($\eta=0.01$) & Adam ($\eta=0.1$) \\[1pt] \hline
training error & $4.00\times 10^{-7}$ & $5.62\times 10^{-4}$ & $4.57\times 10^{-5}$ & $2.02\times 10^{-7}$ & $\mathbf{1.02\times 10^{-7}}$\\[1pt]
test error & $\mathbf{1.38\times 10^{-5}}$ & $8.38\times 10^{-2}$ & $1.19\times 10^{-3}$ & $1.44\times 10^{-5}$ & $1.97\times 10^{-5}$\\[1pt]
\bottomrule
\end{tabular}
\vspace{1mm}
\caption{Training and test errors associated with different optimizers. Training has been stopped after the training time exceeded 60 seconds. SGD and Adam results are based on the same initial neural-network parameters. For EKI, the initial ensemble size is 22 and the diagonal elements of the covariance matrix are updated according to the exponential scheduler (\ref{eq:exp_schedule}).}
\label{tab:pendulum_performance}
\end{table*}
To study the ability of neural ODEs that are trained with EKI to efficiently learn a dynamical system based on given observation data, we first consider the dynamical system
\begin{equation}
    \begin{pmatrix}
    \dot{x}_1 \\
    \dot{x}_2
    \end{pmatrix}
    =
    \begin{pmatrix}
    -0.05 & 1 \\
    -1 & -0.05
    \end{pmatrix}
    \begin{pmatrix}
    x_1 \\
    x_2
    \end{pmatrix}
\label{eq:2d_example}
\end{equation}
with initial condition $(x_1(0),x_2(0))^\top=(1,0)^\top$. The solution of this initial value problem is 
\begin{equation}
    \begin{pmatrix}
    x_1(t) \\
    x_2(t)
    \end{pmatrix}
       =
    \begin{pmatrix}
    e^{-t/20} \cos(t) \\
    -e^{-t/20} \sin(t)
    \end{pmatrix}\,.
\end{equation}

To train a neural ODE $f_\theta(x(t),t)$ to represent the vector field associated with Eq.~(\ref{eq:2d_example}), we use $100$ reference points from a discretized solution of the initial value problem as training data. The discretized solution consists of $500$ time points $t_\ell\in[0,40]$ and the set of $100$ training data points consists of 10 subsets of 10 points associated with consecutive time steps [see red dots in Fig.~\ref{fig:neural_ode_spiral}(a)]. The first points in each subset are selected uniformly at random without replacement from the set of $500$ points.

We represent $f_\theta(x(t),t)$ using a neural network with one hidden layer and 10 $\tanh$ neurons. The total number of parameters is 52. If not stated otherwise, weights and biases are initialized from $\mathcal{U}(-\sqrt{d},\sqrt{d})$, where $d$ is the inverse of the number of input features. In all examples, we integrate neural ODEs using a Dormand--Prince method~\cite{dormand1980family,hairer1993solving}. For EKI, we use an ensemble size of $J=22$ and a diagonal covariance matrix with entries ${(\Gamma_m)_{ij}=\gamma_m \delta_{ij}}$, where $m$ is the current training epoch and $\delta_{ij}$ is the Kronecker delta function (\ie, $\delta_{ij}=1$ if $i=j$ and 0 otherwise). We initially set $\gamma_{0}=0.9$ and then reduce its value every two iterations using an exponential scheduler. That is, every two iterations, we set
\begin{equation}
    \gamma_{m}=\gamma_0 e^{-\alpha m}\,,
    \label{eq:exp_schedule}
\end{equation}
where $\alpha>0$ modulates the decrease of $\gamma_0$. We use an exponential decay in $\gamma_m$ to map small differences between $G(\theta^{(j)}_m)$ and $y$ in Eq.~(\ref{eq:dthetadt}) to noticeable updates in $\theta^{(j)}_m$. In our simulations, we set $\alpha=0.35$. 

Figure~\ref{fig:neural_ode_spiral}(b) shows that this adaptive EKI method is able to train the described neural ODE to represent the dynamical system (\ref{eq:2d_example}) with initial condition ${(x_1(0),x_2(0))^\top=(1,0)^\top}$. Without exponential scheduler, the MSE decreases more slowly as a function of training epochs. To compare the EKI-based solution with a solution that uses a gradient-based optimizer, we train the same neural ODE using the Adam optimizer~\cite{DBLP:journals/corr/KingmaB14}, an adaptive gradient-descent method, with a learning rate of $\eta=0.01$.\footnote{In all numerical experiments that use the Adam optimizer, we set $\beta_1=0.9$, $\beta_2=0.999$, and $\epsilon=10^{-8}$. Here, $\beta_1,\beta_2$ are coefficients that are used to compute running averages of the gradient and its square. The parameter $\epsilon$ is used in the denominator of gradient updates to improve numerical stability.}

For an appropriate comparison of the two optimization methods, we stop the neural ODE training when the training time exceeds one minute on a single core of an Intel\textsuperscript{\textregistered} Core\textsuperscript{\texttrademark} i7-10510U CPU @ 1.80GHz × 8. The total numbers of training epochs of EKI and Adam are 66 (\ie, ca.\ 1 per second) and 2277 (\ie, ca.\ 38 per second), respectively. The training and test errors of EKI are $4.89\times 10^{-7}$ and $9.11\times 10^{-4}$, respectively. The training error of Adam is $1.03\times 10^{-6}$, about twice as large as that of EKI, while the test error of Adam is $1.12\times 10^{-3}$ and thus almost equivalent to that of EKI. We also performed numerical experiments for an additional learning rate ($\eta=0.1$) and for stochastic gradient descent (SGD). The results are summarized in Table~\ref{tab:spiral_performance}. While Adam can achieve a smaller training error for the learning rate $\eta=0.1$, the corresponding test error is substantially larger than for the training with $\eta=0.01$. The performance of SGD is inferior to Adam for the two tested learning rates. Overall, EKI can deliver a competitive performance against the adaptive learning method Adam.

\begin{figure}[ht!]
\vskip 0.2in
\begin{center}
\centerline{\includegraphics{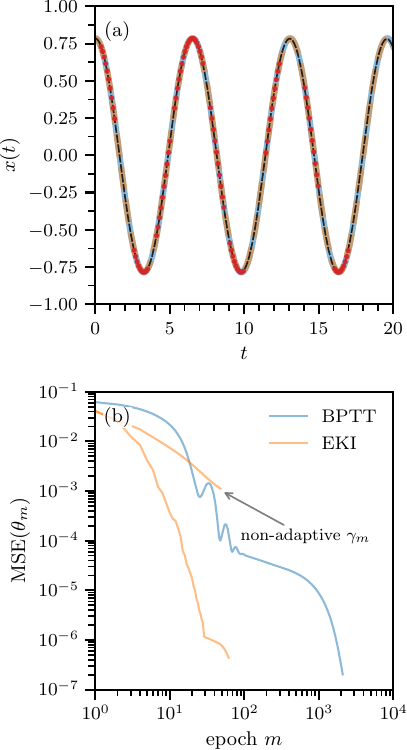}}
\caption{Learning a dynamical system from observation data using BPTT and EKI. (a) We employ a neural ODE to learn the dynamical system (\ref{eq:pendulum}). Solid blue and orange lines correspond to solutions that are obtained with BPTT and EKI, respectively. The dashed black line indicates the solution of the original dynamical system for ${x(0)=\pi/4}$ and ${\dot{x}(0)=0}$. Red dots represent observation data that is used to train the neural ODE. (b) Mean squared error (MSE) associated with BPTT (solid blue line) and EKI (solid orange line) as a function of the number of training epochs $m$ [see Eq.~(\ref{eq:mse_loss})]. The neural ODE consists of one hidden layer with 10 $\tanh$ neurons. Its total number of parameters is 52. In the gradient-based optimization, we use the Adam optimizer and set the learning rate to $\eta=0.01$. The EKI optimizer has 22 ensemble members and an exponential scheduler to adapt $\gamma_m$ [see Eq.~(\ref{eq:exp_schedule})]. For EKI, the MSE is based on the minimum MSE in the whole ensemble. We stopped both training algorithms when their runtime exceeded one minute.}
\label{fig:neural_ode_pendulum}
\end{center}
\vskip -0.2in
\end{figure}
As an example of a non-linear dynamical system, we consider the ODE
\begin{equation}
    \ddot{x}=-\omega \sin(x)\,,
    \label{eq:pendulum}
\end{equation}
which describes the motion of a simple pendulum with natural frequency $\omega$. In our numerical experiments, we set $\omega=1$. As initial condition, we use ${x(0)=\pi/4}$ and ${\dot{x}(0)=0}$. As in the previous example, the neural network parameterizing $f_\theta(x(t),t)$ has 10 $\tanh$ neurons.

To train a neural ODE $f_\theta(x(t),t)$ to represent the vector field associated with Eq.~(\ref{eq:pendulum}), we use $100$ reference points from a discretized solution of the initial value problem as training data. The discretized solution consists of $200$ time points $t_\ell\in[0,20]$ and the set of $100$ training data points consists of 10 subsets of 10 points associated with consecutive time steps [see red dots in Fig.~\ref{fig:neural_ode_pendulum}(a)]. The first points in each subset are selected uniformly at random without replacement from the set of $200$ points.

We use $J=22$ ensemble members and an exponential scheduler (\ref{eq:exp_schedule}) with $\gamma_0=1.4$--$2.6$ and $\alpha=0.4$. The adaptive $\gamma_m$ again helps the EKI-based optimizer reach small loss values [see Fig.~\ref{fig:neural_ode_pendulum}(b)]. We again stop training when the training time exceeds one minute. Figure~\ref{fig:neural_ode_pendulum}(b) shows that the training loss associated with EKI is slightly larger than that associated with Adam ($4.00\times 10^{-7}$ vs.\ $2.02\times 10^{-7}$). We also performed additional numerical experiments using the Adam optimizer with a learning rate of $\eta=0.1$ and SGD with $\eta=0.1,0.01$. The training and test errors are summarized in Table~\ref{tab:pendulum_performance}. The smallest test error has been achieved with EKI.
\section{Optimal control}
\label{sec:optimal_control}
We will now focus on the gradient-free training of neural ODE controllers. As a starting point, we consider the boundary value problem
\begin{equation}
    \dot{x}=f(x(t),u(t),t)\,,\quad {x}(0)={x}_0\,,\quad {x}(T)={x}^*\,,
\label{eq:dyn_sys}
\end{equation}
where the vector field $f\colon \mathbb{R}^n\times \mathbb{R}^m\times \mathbb{R}\rightarrow \mathbb{R}^n$ describes the evolution of the system state $x(t)\in\mathbb{R}^n$ subject to a control function $u(t)\in\mathbb{R}^m$.\footnote{To adhere to the standard notation in control theory, we use $m$ to denote the dimension of the vector space associated with the control signal $u(t)$. Therefore, it is important to distinguish $m$ from the number of epochs.}

In optimal control, one wishes to identify a Lebesgue-measurable control function $u(t)$ that satisfies the constraint (\ref{eq:dyn_sys}) and minimizes the functional
\begin{equation}
    \mathcal{J} = \phi(x(T),T)+\int_0^T L(x(t),u(t),t)\,\mathrm{d}t\,,
\end{equation}
where $\phi\colon \mathbb{R}^n\times \mathbb{R}\rightarrow \mathbb{R}$ and $L\colon \mathbb{R}^n\times \mathbb{R}^m \times \mathbb{R}\rightarrow \mathbb{R}$~\cite{speyer2010primer,lewis2012optimal}. That is, one wishes to minimize $\mathcal{J}$ and steer the dynamical system as defined in Eq.~(\ref{eq:dyn_sys}) from its initial state $x_0\in\mathbb{R}^n$ to a desired target state $x^*\in\mathbb{R}^n$ in finite time $T$. If we were to impose additional constraints on the control function $u(t)$, it has to be chosen from a corresponding set of admissible controls~\cite{wang2018admissible}. 

In this work, we focus on the  special case where $\phi(x(T),T)\equiv 0$ and $L(x(t),u(t),t)\equiv L(u(t))=\|u(t)\|_2^2$. The condition $\phi(x(T),T)\equiv 0$ means that the endpoint cost is zero. Because $L(u(t))=\|u(t)\|_2^2$, the running (or integrated) cost is positive for a non-zero control signal. The corresponding cost function is given by the control energy
\begin{equation}
    E_T[u]=\int_0^T \|u(t)\|_2^2\,\mathrm{d}t
\label{eq:control_energy}
\end{equation}

The outlined optimal control problem aims at finding the control signal $u^*(t)$ that is associated with the smallest control energy and satisfies the constraint (\ref{eq:dyn_sys}). That is,
\begin{equation}
u^*(t)=\mathrm{arg\,min}_{u(t)}E_T[u]\,.
\label{eq:u_min}
\end{equation}
subject to the constraint (\ref{eq:dyn_sys}).

Using Pontryagin's maximum principle~\cite{pontryagin1987mathematical}, a necessary condition for optimal control, we can find the control that satisfies Eqs.~(\ref{eq:dyn_sys}) and (\ref{eq:u_min})  by minimizing the control Hamiltonian
\begin{equation}
H(x(t),u(t),\lambda(t),t)=\lambda(t)^\top f(x(t),u(t),t)+\|u(t)\|_2^2\,,
\label{eq:hamiltonian}
\end{equation}
at every time point $t$. The time-dependent components of the Lagrange multiplier vector $\lambda(t)\in\mathbb{R}^n$ are called the adjoint (or costate) variables of the system~\cite{speyer2010primer,bertsekas2012dynamic}.
\subsection{Neural ODE controllers}
While Pontryagin's maximum principle provides a necessary condition for optimal control~\cite{pontryagin1987mathematical}, the Hamilton--Jacobi--Bellman equation offers both necessary and sufficient conditions for optimality~\cite{zhou1990maximum,bellman2015applied}. Nonlinear optimal control problems are usually solved through indirect and direct numerical methods, and recently transformation methods have been proposed to convert non-linear control problems into linear ones~\cite{kaiser2021data}. Indirect optimal control solvers involve different kinds of shooting methods~\cite{oberle2001bndsco} that use the maximum principle and a control Hamiltonian to construct a system of equations that describe the evolution of state and adjoint variables. On the other hand, direct methods involve parameterizing state and control functions and solving the resulting optimization problem. Possible function parameterizations include piecewise constant functions and other suitable basis functions~\cite{bock1984multiple}. Over the past two decades, pseudospectral methods have been a successful approach to solving nonlinear optimal control problems, with applications in aerospace engineering~\cite{gong2006pseudospectral}. However, it has been shown that certain pseudospectral methods are incapable of solving standard benchmark control problems~\cite{fahroo2008advances}. Here, we parameterize and learn control functions $u(t)$ using neural ODEs $u_{\theta}(t)$ with parameters $\theta\in\mathbb{R}^N$~\cite{asikis2022neural,bottcher2022ai,bottcher2022near}. The boundary value problem (\ref{eq:dyn_sys}) hence becomes
\begin{equation}
    \dot{x}=f(x(t),u_{\theta}(t),t)\,,\quad {x}(0)={x}_0\,,\quad {x}(T)={x}^*\,.
\label{eq:dyn_sys_nn}
\end{equation}

Before applying EKI to optimal control problems, we have to adapt both the underlying inverse problem and ensemble iterations. Recall that the standard EKI algorithm as summarized in Section~\ref{sec:eki} aims at finding a solution to the inverse problem (\ref{eq:inverse_problem}) by minimizing the functional
\begin{equation}
    \Phi(\theta;y)=\frac{1}{2}\|y-G(\theta)\|_{\Gamma}^2\equiv \frac{1}{2}\|\Gamma^{-1/2}(y-G(\theta))\|_{\mathcal{Y}}^2\,.
\end{equation}
To account for the additional regularization term in optimal control problems, we extend the inverse problem (\ref{eq:inverse_problem}) using
\begin{equation}
    z = F(\theta) + \xi\,,
    \label{eq:extended_inverse_problem}
\end{equation}
where $F\colon \mathcal{U}\times \mathcal{U} \rightarrow \mathcal{Z}$, $z=(y,0)^\top\in\mathcal{Z}$, $\xi=(\xi_1,\xi_2)^\top$, $\xi\sim \mathcal{N}(0,\Sigma)$ and
\begin{equation}
    \Sigma =
    \begin{pmatrix}
        \Gamma & 0 \\
        0 & \mu^{-1} \Gamma'
    \end{pmatrix}\,.
\end{equation}
The first element of the function $F(\theta)=\left(G(\theta),H(\theta)\right)^\top$, $G(\theta)$, describes the evolution of the controlled system state $x(t)$ associated with $f(x(t),u_\theta(t),t)$ while the second element, $H(\theta)$, accounts for the control energy term $E_T[u_\theta]$.

As in the unregularized EKI method that we described in Section~\ref{sec:eki}, the evolution of the ensemble $\{\theta^{(j)}\}_{j=1}^J$ associated with the inverse problem (\ref{eq:extended_inverse_problem}) is given by
\begin{align}
    \dot{\theta}^{(j)}&=-B^{\theta F}(\theta) \Sigma^{-1} \left(F(\theta^{(j)})-z\right)\,,\label{eq:dthetadt_regularized}\\
    \theta^{(j)}(0)&=\theta_0^{(j)}\,,
\end{align}
where the empirical cross-covariance matrix, $B^{\theta F}(\theta)$, and the ensemble mean, $\bar{F}$, are given by
\begin{equation}
    B^{\theta F}(\theta)=\frac{1}{J}\sum_{j=1}^J \left(\theta^{(j)}-\bar{\theta}\right)\otimes \left(F(\theta^{(j)})-\bar{F}\right)\,.
    \label{eq:b_ftheta}
\end{equation}
and
\begin{equation}
    \bar{F}=\frac{1}{J}\sum_{j=1}^J F(\theta^{(j)})\,,
    \label{eq:bar_f}
\end{equation}
respectively. The associated loss function is 
\begin{equation}
    \tilde{\Phi}(\theta;z)=\frac{1}{2}\|\Sigma^{-1/2}(z-F(\theta))\|_{\mathcal{Z}}^2\,.
\end{equation}
We now identify $y$ in $z=(y,0)^\top$ with the target state $x^*$, and we set $G(\theta)=x(T;\theta)$ and $H(\theta)=E_T[u_\theta]^{1/2}$. This yields the loss function
\begin{equation}
    \tilde{\Phi}(\theta;x^*)=\frac{1}{2}\|x(T;\theta)-x^*\|_\Gamma^2+\frac{\mu}{2\Gamma'} E_T[u_\theta]\,.
    \label{eq:eki_control_loss}
\end{equation}
The above formulation of optimal control problems is similar in its mathematical structure to Tikhonov EKI that has been introduced by~\citet{chada2020tikhonov} to regularize the parameter vector $\theta$ while solving inverse problems using EKI.
\subsection{Numerical experiments}
\begin{figure*}[ht!]
\vskip 0.2in
\begin{center}
\centerline{\includegraphics{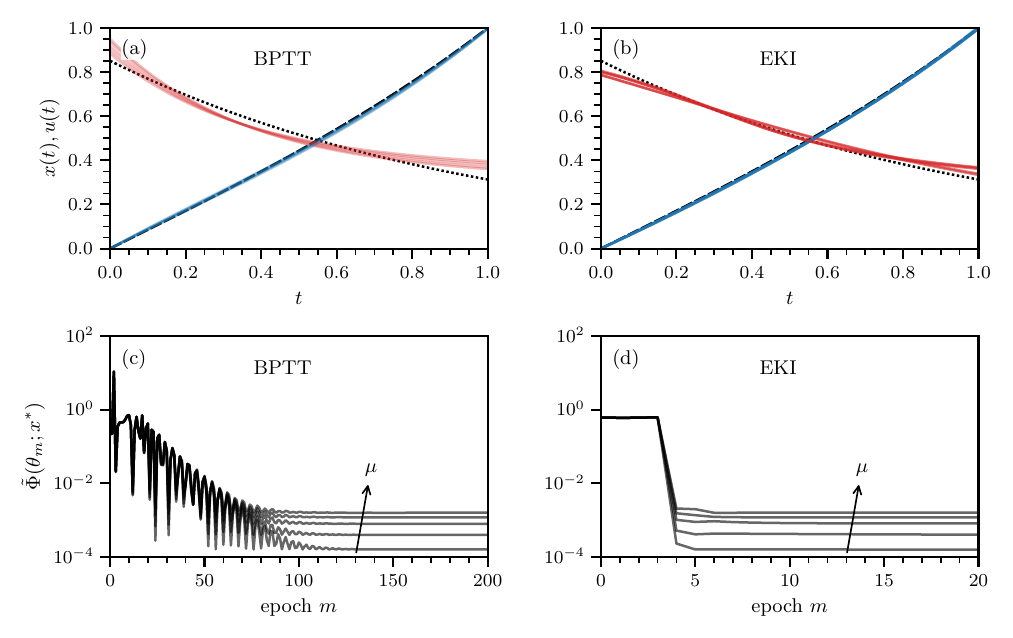}}
    \caption{Controlling linear dynamics with neural ODEs. The neural ODE controller $u_\theta(t)$ has been trained with BPTT and EKI in panels (a,c) and (b,d), respectively. (a,b) Evolution of the system state $x(t)$ and control function $u(t)$. Dashed and dotted lines show solutions $x^*(t)=\sinh(t)/\sinh(1)$ and $u^*(t)=\exp(-t)/\sinh(1)$ associated with the optimal control problem (\ref{eq:energy_constraint}) and (\ref{eq:dyn_sys_simple_time}). We set $x_0=0$ and $T=a=b=x^*=1$ in all simulations. Numerical solutions in panels (a,b) are indicated by colored lines. Different lines correspond to different values of the control energy regularization parameter $\mu\in\{0.001,0.0025,0.005,0.0075,0.01\}$. The neural ODE controller $u_\theta(t)$ consists of three hidden layers with five exponential linear units (ELU) neurons each. Its total number of parameters is 106. In the gradient-based optimization, we used the Adam optimizer and set the learning rate to $\eta=0.175$. (c,d) The loss function $\tilde{\Phi}(\theta_m;x^*)$ as a function of the number of training epochs $m$. For EKI, we set $\Gamma=0.3,\Gamma'=0.01$ and show the minimum loss in the whole ensemble. After three iterations, we set $\Gamma=0.15$. For BPTT, ensemble variances are not relevant and we thus optimize the loss function for which $\Gamma=\Gamma'=1$.}
    \label{fig:bptt_vs_eki}
\end{center}
\vskip -0.2in
\end{figure*}
We now examine if neural ODE controllers that are trained with EKI are able to learn effective control functions. To do so, we consider the optimal control problem that aims at identifying
\begin{equation}
u^*(t)=\mathrm{arg\,min}_{{u}(t)}E_T[{u}]
\label{eq:energy_constraint}
\end{equation}
subject to
\begin{equation}
\dot{x}=ax+bu\,,\quad {x}(0)={x}_0\,,\quad {x}(T)={x}^*\,.
\label{eq:dyn_sys_simple_time}
\end{equation}
The mathematical structure of this optimal control problem is similar to that of control problems encountered in regulating temperature within a room~\cite{lewis2012optimal}. 

We use Pontryagin's maximum principle to derive $u^*(t)$, which we will assume to be the optimal control signal associated with Eqs.~(\ref{eq:energy_constraint}) and (\ref{eq:dyn_sys_simple_time}). This calculation yields
\begin{equation}
u^*(t)=\frac{a e^{-a t}}{b \sinh(a T)} \left(x^*-x_0e^{a T}\right)\,.
\label{eq:time_dependent_control}
\end{equation}
For $a>0$, the magnitude of $u^*(t)$ decays exponentially with $t$ because the value of the system state of the uncontrolled dynamics $\dot{x}=a x$ can be influenced more effectively for small values of $t$ than for large ones. The opposite holds for $a<0$. 

The system state $x(t)$ under the influence of $u^*(t)$ is
\begin{equation}
x^*(t)=x_0 e^{a t}+\frac{\sinh (a t)}{\sinh (a T)} \left(x^*-x_0 e^{a T}\right)\,.
\end{equation}
Finally, the control energy associated with the optimal control signal (\ref{eq:time_dependent_control}) is
\begin{equation}
E_T[u^*]=\int_0^T {u^*(t)}^2\,\mathrm{d}t=\frac{a(1-e^{-2 a T})(x^*-x_0 e^{a T})^2}{2 b^2 \sinh^2(a T)}\,.
\label{eq:control_energy_time_dependent}
\end{equation}

In the following numerical experiments, we consider the optimal control problem (\ref{eq:energy_constraint}) and (\ref{eq:dyn_sys_simple_time}) for which $x_0=0$ and $T=a=b=x^*=1$. The corresponding optimal control signal, system state, and control energy are $u^*(t)=\exp(-t)/\sinh(1)$, $x^*(t)=\sinh(t)/\sinh(1)$, and $E_T[u^*]=2/(e^2-1)\approx 0.313$. The neural ODE controller $u_\theta(t)$ that we employ consists of three hidden layers with five exponential linear units (ELU) neurons each. Its total number of parameters is 106. 

In the gradient-based optimization (\ie, for BPTT), we do not use an ensemble of ANN parameters and thus minimize the loss function \eqref{eq:eki_control_loss} for which $\Gamma=\Gamma'=1$. We use the Adam optimizer and set the learning rate to $\eta=0.175$. We evaluated different learning rates within the range of 0.1 to 0.2 and chose the learning rate that produced solutions that closely resembled the optimal ones. 

For EKI, the initial number of ensemble members is ${J=2}$ and we set $\Gamma=0.3,\Gamma'=0.01$. After three iterations, we add 20 new ensemble members and set $\Gamma=0.15$. We observe that a small number of ensemble members yields a good initial performance. Adding new ensemble members helps EKI learn good parameters as the optimization progresses. These observations on expanding the initial ensemble are consistent with related work~\cite{kovachki2019ensemble}.

To investigate how the control energy regularization parameter $\mu$ [see Eq.~(\ref{eq:eki_control_loss})] affects the learned system state and control function, we vary $\mu$ within the range of 0.001 to 0.01 in the loss function $\tilde{\Phi}(\theta;x^*)$. The dashed and dotted black lines in Fig.~\ref{fig:bptt_vs_eki}(a,b) represent the optimal solutions $x^*(t)$ and $u^*(t)$. Solid red lines show the control solutions $u_\theta(t)$ learned by neural ODE controllers. Solid blue lines indicate the corresponding system state evolution. As we show in Fig.~\ref{fig:bptt_vs_eki}(a,b), variations in $\mu$ produce variations in the learned solutions of the optimal control problem. Furthermore, the shown simulation results suggest that both the gradient-based optimizer and EKI are capable of training neural ODE controllers to learn control functions that closely approximate the optimal solution. 

We calculated the MSE between the learned and optimal control solution to compare the quality of the solutions learned using BPTT and EKI. We find the the MSE is about $1$--$1.4\times 10^{-3}$ (BPTT) and $0.4$--$0.6\times 10^{-3}$ (EKI). The MSE associated with EKI is thus about 50\% of that associated with the BPTT-based solution.

Figure~\ref{fig:bptt_vs_eki}(c,d) shows the loss function $\tilde{\Phi}(\theta_m;x^*)$ associated with BPTT and EKI as a function of the number of training epochs $m$. Different loss trajectories correspond to different control energy regularization parameters $\mu$. After approximately 90-100 epochs, Adam-based optimization stabilizes at certain loss values, although it may achieve small loss values earlier in the training process. In contrast, EKI is able to achieve similar results within only 4-5 epochs. In the studied control example, EKI and Adam-based optimization achieve about 1 and 10 optimization steps per second.

Our results show that EKI can account for a control energy regularization term and be employed to solve an optimal control problem in a gradient-free manner.
\section{Discussion and conclusion}
We studied the ability of ensemble Kalman inversion (EKI) as an alternative method to backpropagation for training neural ODEs in system identification and optimal control tasks. Ensemble Kalman inversion is a gradient-free optimization method that solves general inverse problems within a Bayesian framework. It only requires one to evaluate artificial neural networks in forward passes, making backward passes unnecessary. 

After providing an overview of the basic formalism of EKI, we applied this method to system identification problems associated with linear and non-linear dynamics. Our results showed that EKI performs well with respect to gradient-based optimization methods such as SGD and Adam. We also applied EKI to optimal control problems that involve an additional control energy regularization term to keep the integrated square norm of the control signal small [see Eqs.~(\ref{eq:control_energy}) and (\ref{eq:eki_control_loss})]. We reformulated EKI iterations to account for such a regularization term and applied this adapted method to an optimal control problem with an underlying linear vector field. The EKI approach that we use for solving optimal control problems is similar in its mathematical structure to Tikhonov EKI~\cite{chada2020tikhonov}.

In summary, our results suggest that EKI can serve as an effective alternative to gradient-based optimization techniques in training neural ODEs for system identification problems. Additionally, we extended the use of EKI to optimal control problems, providing a new perspective on solving inverse problems that arise in control theory using EKI.

There are several promising avenues for future research. For instance, it would be interesting to examine EKI's effectiveness in higher-dimensional system identification and control tasks, and contrast its performance with other gradient-based techniques that have been utilized in training neural ODEs~\cite{chen2018neural,ainsworth2021faster}. Moreover, future research may study the geometric properties of the optima found by gradient-based methods and EKI, providing insights into the similarities and differences between these optimization approaches. Such an analysis can broaden our understanding of optimization landscapes~\cite{bottcher2022visualizing} and help guide the development of optimization algorithms that are both robust and efficient. Additionally, future work may focus on EKI's capacity to train neural ODEs using noisy observation data or other types of observation data that make the use of gradient-based methods more challenging. Finally, investigating the use of EKI in training physics-informed neural networks and their corresponding controllers~\cite{mowlavi2022optimal} would be a valuable area of investigation as well.
\bibliography{refs}
\bibliographystyle{icml2023}

%

\end{document}